\documentclass[journal]{IEEEtai}

\usepackage[colorlinks,urlcolor=blue,linkcolor=blue,citecolor=blue]{hyperref}

\usepackage{color,array}

\usepackage{graphicx}
\usepackage{scalerel}
\usepackage{tikz}
\usepackage[super]{nth}
\usetikzlibrary{svg.path}

\definecolor{orcidlogocol}{HTML}{A6CE39}
\tikzset{
  orcidlogo/.pic={
    \fill[orcidlogocol] svg{M256,128c0,70.7-57.3,128-128,128C57.3,256,0,198.7,0,128C0,57.3,57.3,0,128,0C198.7,0,256,57.3,256,128z};
    \fill[white] svg{M86.3,186.2H70.9V79.1h15.4v48.4V186.2z}
                 svg{M108.9,79.1h41.6c39.6,0,57,28.3,57,53.6c0,27.5-21.5,53.6-56.8,53.6h-41.8V79.1z M124.3,172.4h24.5c34.9,0,42.9-26.5,42.9-39.7c0-21.5-13.7-39.7-43.7-39.7h-23.7V172.4z}
                 svg{M88.7,56.8c0,5.5-4.5,10.1-10.1,10.1c-5.6,0-10.1-4.6-10.1-10.1c0-5.6,4.5-10.1,10.1-10.1C84.2,46.7,88.7,51.3,88.7,56.8z};
  }
}

\newcommand\orcid[1]{\href{https://orcid.org/#1}{\mbox{\scalerel*{
\begin{tikzpicture}[yscale=-1,transform shape]
\pic{orcidlogo};
\end{tikzpicture}
}{|}}}}

\usepackage{hyperref} 
\hypersetup{hidelinks}
\usepackage{multicol}
\usepackage{multirow}

\ifCLASSOPTIONcompsoc
  \usepackage[nocompress]{cite}
\else
  \usepackage{cite}
\fi

\usepackage{graphicx}

\usepackage{amsmath}
\usepackage{caption}
\usepackage{subcaption}

\usepackage{framed,multirow}
\usepackage[font=small,labelfont=bf]{caption}
\usepackage[T1]{fontenc}
\usepackage[utf8x]{inputenc} 
\usepackage{babel}
\usepackage{xcolor}
\usepackage[table]{colortbl}
\usepackage{collcell}
\usepackage{hhline}
\usepackage{pgf}
\usepackage{multirow}

\def\colorModel{hsb} 

\newcommand\ColCell[1]{
  \pgfmathparse{#1<25?1:0}  
    \ifnum\pgfmathresult=0\relax\color{white}\fi
  \pgfmathsetmacro\compA{0}      
  \pgfmathsetmacro\compB{#1/50} 
  \pgfmathsetmacro\compC{1}      
  \edef\x{\noexpand\centering\noexpand\cellcolor[\colorModel]{\compA,\compB,\compC}}\x #1
  } 
\newcolumntype{E}{>{\collectcell\ColCell}m{0.5cm}<{\endcollectcell}}  
\newcommand*\rot{\rotatebox{90}}
\usepackage{pifont}
\usepackage{color}
\usepackage{alltt}

\usepackage{enumerate}
\usepackage{siunitx}
\usepackage{breakurl}
\usepackage{epstopdf}
\usepackage{pbox}
\usepackage[export]{adjustbox}

\usepackage{algorithmic}
\usepackage{algorithm}
\usepackage{url}
\usepackage{booktabs}

\begin{document}

\title{Landmark-Aware and Part-based Ensemble Transfer Learning Network for Facial Expression Recognition from Static images} 

\author{Rohan Wadhawan \orcid{0000-0001-8100-668X},~\IEEEmembership{Member,~IEEE,}
        Tapan K. Gandhi \orcid{0000-0002-3532-9389},~\IEEEmembership{Senior Member, ~IEEE}
\thanks{R. Wadhawan and T.K.Gandhi are with the Department of Electrical Engineering, Indian Institute of Technology-Delhi, New Delhi 110016, India. \\
E-mail: rohanwadhawan7@gmail.com and tgandhi@ee.iitd.ac.in}
}
 


\maketitle

\begin{abstract}
Facial Expression Recognition from static images is a challenging problem in computer vision applications. Convolutional Neural Network (CNN), the state-of-the-art method for various computer vision tasks, has had limited success in predicting expressions from faces having extreme poses, illumination, and occlusion conditions. To mitigate this issue, CNNs are often accompanied by techniques like transfer, multi-task, or ensemble learning that often provide high accuracy at the cost of increased computational complexity. In this work, we propose a Part-based Ensemble Transfer Learning network that models how humans recognize facial expressions by correlating the spatial orientation pattern of the facial features with a specific expression. It consists of 5 sub-networks, and each sub-network performs transfer learning from one of the five subsets of facial landmarks: eyebrows, eyes, nose, mouth, or jaw to expression classification. We show that our proposed ensemble network uses visual patterns emanating from facial muscles' motor movements to predict expressions and demonstrate the usefulness of transfer learning from Facial Landmark Localization to Facial Expression Recognition. We test the proposed network on the CK+, JAFFE, and SFEW datasets, and it outperforms the benchmark for CK+ and JAFFE datasets by 0.51\% and 5.34\%, respectively. Additionally, the proposed ensemble network consists of only 1.65M model parameters, ensuring computational efficiency during training and real-time deployment. Grad-CAM visualizations of our proposed ensemble highlight the complementary nature of its sub-networks, a key design parameter of an effective ensemble network. Lastly, cross-dataset evaluation results reveal that our proposed ensemble has a high generalization capacity, making it suitable for real-world usage.
\end{abstract}


\section{Introduction}

\IEEEPARstart {F}{acial Expressions} play a central role in human-to-human interactions. Humans recognize expressions quickly, even under challenging conditions like poor illumination, occlusion, and non-frontal poses. On the other hand, machine interpretation of human faces is still evolving. Modern-day computers use machine learning techniques, like neural networks, to improve human-computer interaction through automated Face Detection (FD), Facial Landmark Localization (FLL), Facial Recognition (FR), and Static image or Dynamic Facial Expression Recognition. From this point onwards, we will refer to Static Image Facial Expression Recognition by the abbreviation FER.

Over the years, a large number of datasets have become available for training, evaluating, and benchmarking FER techniques, like the Extended Cohn-Kanade (CK+) \cite{ckplus-dataset-paper}, Japanese Female Facial Expression (JAFFE) \cite{jaffe-dataset-paper}, Static Facial Expressions in the Wild (SFEW 2.0) \cite{sfew-dataset-paper, sfew-dataset-paper-2}, Facial Expression Recognition 2013 (FER2013) \cite{fer-dataset-paper}, and Real-world Affective Faces Database (RAF-DB) \cite{rafdb-dataset-paper}. Further, the intervention of neural networks like the Convolutional Neural Networks (CNN) has significantly improved the state-of-the-art of FER. However, vanilla CNN models have had limited success in analyzing expressions from face images in the challenging conditions mentioned above. To overcome this limitation, techniques like transfer, ensemble, and multitask learning have been employed alongside CNN models to achieve high expression classification accuracy.

Transfer learning (TL) techniques, which usually involve pre-training and fine-tuning, have proven to alleviate overfitting, boost accuracy and reduce computational resources required for training a model \cite{tls-paper}. For example, feature-based TL techniques have been employed for FER, where FER is the target task, and the source task is usually image classification or face recognition. Further, some techniques use a multi-stage fine-tuning process over the direct use of these pre-trained and fine-tuned models \cite{dsp-63}. 

Network Ensemble (here, it refers to the ensemble at the decision level) has outperformed individual networks to achieve high FER accuracy. A good ensemble network has complementary sub-networks, and a suitable decision policy to predict the output \cite{dsp-57, sfew-dsp-4, sfew-dsp-5,dsp-172}. This technique, too, has its shortfall of being computationally expensive, both in time and storage, and is often challenging to implement. It is also vital to train complementary models. Otherwise, an ensemble of similar models may overfit unseen test data.

Multi-task FER networks are jointly trained networks. One network focuses on FER, which is the primary task. The other network(s) perform auxiliary task(s) like AU detection, facial landmarks localization, or face recognition, which helps transfer knowledge from these secondary tasks onto the primary task \cite{dsp-58,dsp-61,dsp-175, dsp-176}. Nevertheless, this technique requires that labeled data be available for each secondary task, and the complexity of implementation increases with the number of secondary tasks.

{\color{black}In this paper, we synthesize the inspiration of how humans recognize expressions and the technical motivation to develop a neural architecture that is accurate, computationally efficient, and robust to challenging scenarios into a novel technique for Facial Expression Recognition. We know that human beings correlate patterns of motor movements in the facial muscles with facial expressions. From a static standpoint, we correlate the spatial patterns of facial features like eyebrows, eyes, nose, mouth, and jaw with a corresponding emotion. This means that recognizing facial expressions depends upon how well an individual can detect and understand these patterns. We translate this observation to the technical domain by drawing an analogy between detecting spatial patterns of facial features and FLL that locates a fixed number of fiducial points on the face. As the change in expression, head pose, or illumination influences these points' relative position, a good FLL model must learn a robust facial representation under these conditions for accurately determining these points. Thus, we fine-tune an FLL model and utilize this robust representation for FER. However, transfer learning from a model that captures the spatial orientation of all facial features at once may lead to misclassification in occluded faces. To overcome this issue, we design a part-based ensemble such that each sub-network focuses on a different facial feature and makes its independent prediction. The sub-networks are modeled to be complementary to reduce the chance of overfitting on the training dataset. Furthermore, we wanted to balance the trade-off between the model's computational complexity (FLOPs for inference and storage requirement) and its accuracy. So, each sub-network is an end-to-end trainable deep neural network with only 24 layers and 0.33M parameters. We make the following contributions in this article :

\begin{itemize}

\item  We propose an ensemble Part-based Transfer Learning network and evaluate it on three datasets: JAFFE, CK+, and SFEW. Here, part refers to specific facial features, like eyebrows, eyes, nose, mouth, and jaw. In addition to our proposed network, we employ two baseline networks that help us evaluate the benefit of transfer learning from facial landmark localization and the merit of a part-based ensemble transfer learning approach.

\item We evaluate the expression classification performance of our proposed network in terms of accuracy, FLOPs for inference, and storage requirement. We compare the accuracy and parameter count of our model with the current state-of-the-art models for these datasets.

\item  We perform a study to verify the usefulness of transfer learning from the Facial Landmark Localization (FLL) task to Facial Expression Recognition task and demonstrate how the facial representations learned from the former can enhance the latter's performance.

\item  We also perform a visual dissection study for which we employ Gradient-weighted Class Activation Mapping (Grad-CAM) \cite{vis-gradcam}. It helps visualize which regions of the face each sub-network network focuses on while performing FER. We also contrast the Grad-CAM visualization of the proposed ensemble network with the baseline networks.

\item We conduct a cross-dataset generalization test for expression classification. We use this test to estimate a model's prediction ability when trained and tested on two datasets with different data characteristics (like gender and ethnicity for facial data) and expression-wise distribution. We select SFEW as the training set and CK+ and JAFFE as the test set for the cross-dataset generalization test. 
   
\end{itemize}

The rest of this article is structured as follows. The datasets used for experiments are presented in Section 2. Our FER deep learning pipeline is described in detail in section 3. The results are presented in section 4. Discussion on the results is provided in section 5. Finally, conclusions and future directions are provided in section 6.
}

\section{Datasets}
We use three datasets to evaluate our proposed methodology, the Extended CohnKanade (CK+) \cite{ckplus-dataset-paper}, Japanese Female Facial Expression (JAFFE) \cite{jaffe-dataset-paper}, and Static Expressions in the Wild (SFEW) \cite{sfew-dataset-paper,sfew-dataset-paper-2}.

\begin{figure}[t]
\centering
\includegraphics[width=7cm,height=1.5cm]{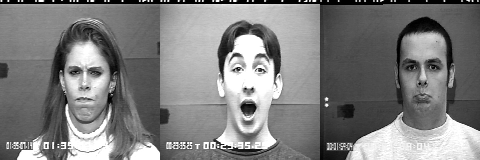}
\caption{Three original photos taken from the CK+ dataset.}
\label{fig:combo_ckplus_dataset}
\vspace{-2.5mm}
\end{figure}

\textbf{CK+ dataset} is the most extensively used laboratory-controlled dataset for evaluating FER systems (some examples are shown in Fig.~\ref{fig:combo_ckplus_dataset}). CK+ contains 593 video sequences from 123 subjects. Each video sequence shows a shift from a neutral facial expression to a peak expression. Among these videos, 327 sequences from 118 subjects are labeled with seven basic expression labels (anger, contempt, disgust, fear, happiness, sadness, and surprise) based on the Facial Action Coding System (FACS). {\color{black} We follow the protocol that uses each sequence's first frame as a neutral frame and the last three frames as expressive frames, thereby obtaining a dataset of 1308 images consisting of 8 expressions (7 basic + neutral) classes in total \cite{ckplus-dsp-1,ckplus-dsp-2,ckplus-dsp-3,ckplus-dsp-4,ckplus-dsp-5}. We adopt a 10-fold subject-independent cross-validation strategy\cite{ckplus-dsp-2,ckplus-dsp-4,ckplus-dsp-5}. In this strategy, we generate ten data folds by sampling subject IDs' in ascending order, such that all images of one subject are in one fold, thereby making each fold subject exclusive of the other. Then, we train on nine of these ten folds and test on the remaining one, such that each fold acts as a test set once. The subject-independent cross-validation helps determine the generalizability of our network to novel subjects.}

\begin{figure}[t]
\centering
\includegraphics[width=7cm,height=1.5cm]{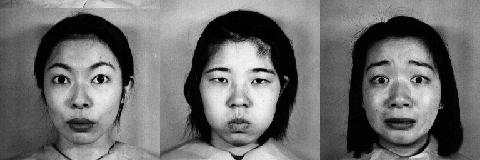}
\caption{Three original photos taken from the JAFFE dataset.}
\label{fig:combo_jaffe_dataset}
\end{figure}

\textbf{JAFFE dataset} is a laboratory-controlled image dataset containing 213 samples of posed expressions from 10 Japanese females (some examples are shown in Fig.~\ref{fig:combo_jaffe_dataset}). Each person has 3 to 4 images across seven facial expressions: anger, disgust, fear, happiness, sadness, surprise, and neutral. The dataset is challenging because it contains a few examples per subject/expression. {\color{black}We use the leave-one-subject-out evaluation strategy \cite{jaffe-1,jaffe-1-1,jaffe-1-2, jaffe-1-3,jaffe-1-4,ckplus-dsp-1} to test our network's generalizability to novel subjects. In this ten-fold subject-independent strategy, we train a model on the images of nine subjects and test it on the tenth subject, such that each subject acts as a test set once. The average performance across the ten models is a good indicator of the proposed method's expression classification ability.}

\begin{figure}[t]
\centering
\includegraphics[width=7cm,height=1.5cm]{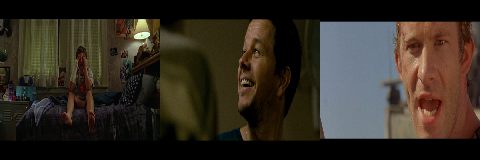}
\caption{Three original photos taken from the SFEW dataset.}
\label{fig:combo_sfew_dataset}
\vspace{-2.5mm}
\end{figure}

\textbf{SFEW 2.0 dataset} is the most widely used benchmark dataset for facial expression recognition in the wild. It provides 1,766 images, comprising 958
train, 436 validation, and 372 test images. Each image belongs to one of the seven expression classes, i.e., anger, disgust, fear, neutral, happy, sad, and surprise. These images have varied head pose (yaw, bobble, and pitch), facial sizes, and contrast, as shown in Fig.~\ref{fig:combo_sfew_dataset}. Moreover, the dataset curators provide train and validation sets' expression labels while holding the test set labels back for the EmotiW 2015 challenge \cite{emotiw2015}. Thus, we report our performance on the validation set for comparison with other methods \cite{ckplus-dsp-5,ckplus-dsp-2,sfew-dsp-6,pw-dsp-15,sfew-dsp-7}. 

\begin{figure}[t]
\centering
\includegraphics[width=7.5cm,height=2.5cm]{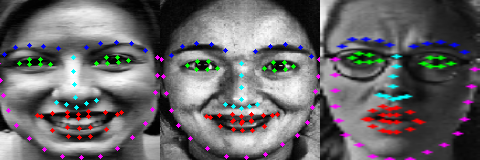}
\caption{68 facial landmark points predicted on facial crops obtained from CK+, JAFFE, and SFEW dataset, in order left-to-right, respectively, using SSD face detection and ERT face landmark localization. Here points corresponding to eyebrows are represented by blue, eyes by green, nose by light blue, mouth by red, and jaw by pink color.}
\label{fig:combo_all_lr}
\end{figure}

\section{Methodology}
\begin{figure*}[h]
\centering
\includegraphics[width=15cm,height=7cm]{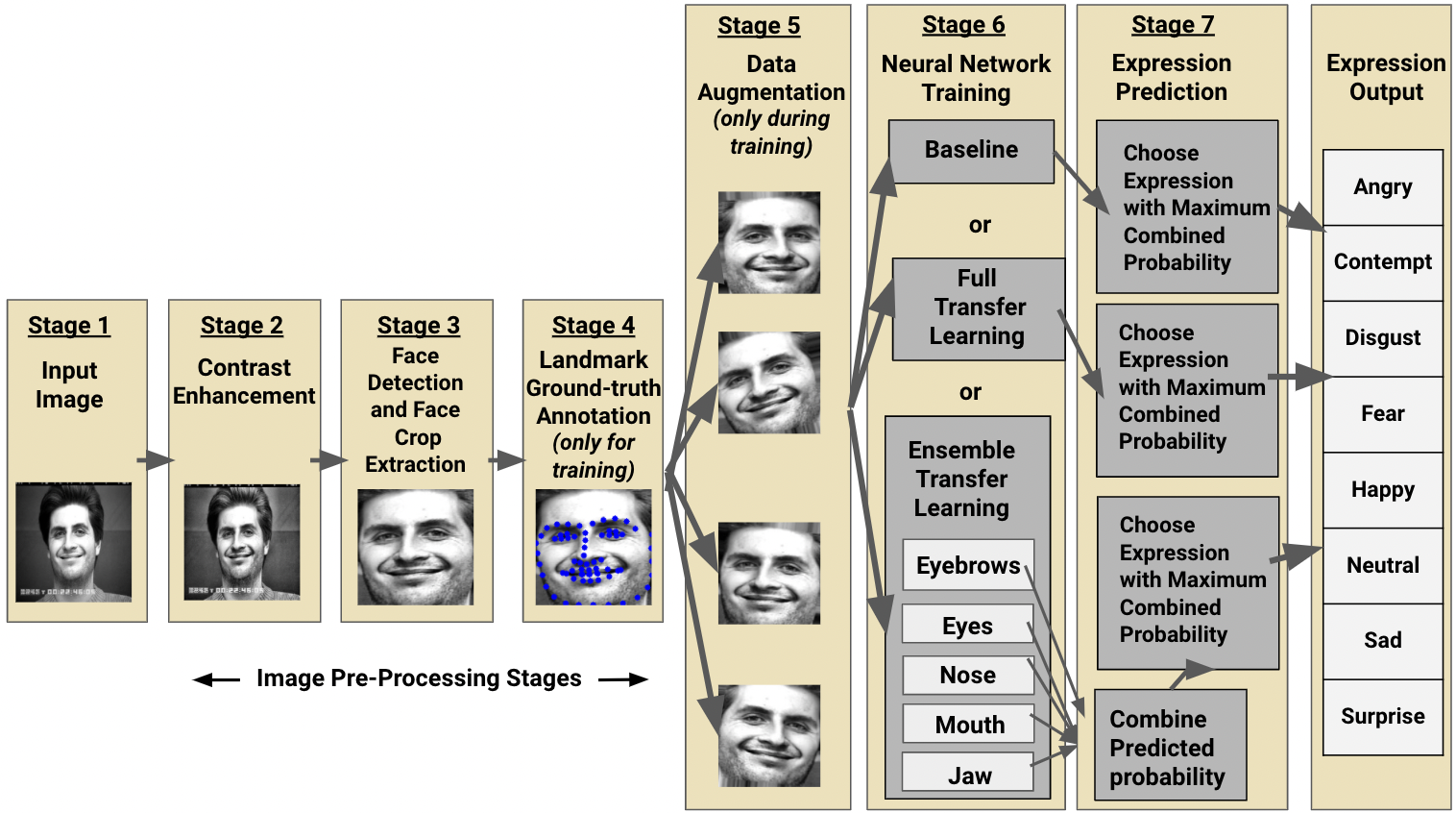}
\caption{The Seven Stages of our Facial Expression Recognition Pipeline. The landmark ground-truth annotation and data augmentation stages are used only during training. Here, the expression \emph{Contempt} is applicable only for the CK+ dataset. }
\label{fig:pipeline_diagram}
\end{figure*}

{\color{black}
Our proposed ensemble network performs transfer learning from the task of Facial Landmark Localization (FLL) to Facial Expression Recognition (FER). We track 68 fiducial points on the face and divide them into five subsets, each corresponding to a specific facial feature - eyebrows, eyes, nose, mouth, or jaw, as shown in Fig. \ref{fig:combo_all_lr}. The fiducial points on the face do not clearly distinguish between jaw, chin, and cheek area, and we represent the three regions together as part of the jaw subset. We train 5 FLL networks, one for each subset of fiducial points. Each FLL network consists of a CNN feature extractor base followed by a landmarks localization head. We perform transfer learning from FLL to FER for each sub-network by replacing the localization head with an expression classification head and fine-tuning it with the shared CNN feature extractor. Finally, we combine these five sub-networks with our ensemble decision policy (Eq. \ref{eq:inference}) and predict the expression class. 

The proposed approach is different from \cite{ckplus-dsp-5}, which performs transfer learning from the task of face recognition to the task of expression recognition. Notable issues with transfer learning from face recognition include using features learned from facial data that primarily exhibit neutral or happy expressions and features containing subject-specific information, which reduce a network's capacity to learn other expressions and are irrelevant for expression recognition tasks respectively\cite{dsp-paper, ckplus-dsp-5}. On the other hand, the FLL task aims to precisely locate facial features across subjects, expressions, poses, and illuminations, thereby enabling it to learn generalized spatial representations of the face. We verify the usefulness of transfer learning from FLL to FER, which is the premise behind our network design and training strategy (Section \ref{sec:relearn}). Further, the proposed technique is different from \cite{dsp-63, ckplus-dsp-5} as we do not pre-train on external data. Unlike \cite{dsp-58}, we use an ensemble network and independently train the landmark output head and classification output head. We train our proposed ensemble using a three-stage training process, described in Section \ref{sec:classes}. Moreover, each sub-network of our proposed ensemble receives the entire face image as input rather than a face patch that contains a specific facial feature. The primary reason for this design is to enable each sub-network to learn facial feature-specific representations in the context of other facial features \cite{sinha2002qualitative}. Unlike patch-based networks, our proposed network avoids the additional computation overhead of extracting image patches corresponding to different facial features. Besides, our method contrasts the ensemble techniques used in \cite{dsp-172,sfew-dsp-4,sfew-dsp-5} as we neither apply a hierarchical committee to ensemble individual networks nor we weigh each sub-network differently. 

Furthermore, we validate our approach against two baseline networks. The first network is named the Baseline network. We do not perform transfer learning for this network and directly train it for expression classification. The purpose of this network is to determine the merit of transfer learning from FLL. The second network is named the Full Transfer Learning (FTL) network. We perform transfer learning from FLL to FER for this network. The purpose of this network is to determine the merit of part-based transfer learning.

The proposed network is part of the facial expression recognition pipeline we have developed. Our pipeline consists of seven stages: Input, Image Contrast Enhancement, Face Detection, Landmark Ground-truth Annotation, Data Augmentation, Neural Network Training, and Expression Prediction, as shown in Fig.~\ref{fig:pipeline_diagram}. Each stage is described in detail in the following subsections.
}

\subsection{Input Image}
We use grayscale images in all the experiments. The grayscale channel is repeated thrice for each image and concatenated to produce a three-channel image as required further down the pipeline. Then, we resize each image to 300x300x3 using bilinear interpolation before feeding it to the face detector (the number of channels has been empirically determined). Any subsequent rescaling of the image also uses bilinear interpolation for consistency.

\subsection{Image Pre-Processing}
Image Pre-Processing accounts for three stages in the pipeline: Image Contrast Enhancement, Face Detection, and Landmark Ground-truth Annotation. We use Contrast Limited Adaptive Histogram Equalization (CLAHE) \cite{clahe} to perform contrast enhancement of all images across all three datasets. After contrast enhancement, we apply Opencv's  SSD detector \cite{fd-ssd, opencv_library} on the images and obtain facial crops from it based on the bounding box predicted by the detector. We resize each face crop to the size 160x160x3, as this is the input size expected by the neural network. Then, we move to the Facial  Landmark Ground-truth Annotation Stage. Our approach performs transfer learning from the task of FLL to expression classification. FLL training gives rise to the need for annotated data for facial landmarks in addition to expression labels. However, FER datasets either do not have landmark annotations or differ in the number of landmarks annotated, method of annotation (manual or automated), and the choice of automated annotation algorithm. CK+ dataset provides 68 landmark positions for each image by using Active Appearance Model (AAM) tracking technique \cite{lr-aam-org,lr-aam-graddesc}. JAFFE dataset does not provide landmark annotation. Lastly, the SFEW dataset uses a different AAM implementation \cite{lr-aam-sfew} for annotating the landmark positions. Moreover, the number of landmark positions differs among SFEW images; some have 39, and others have 68 positions annotated. Therefore, we need a uniform method to annotate ground truth landmark positions across all datasets. Further, we do not intend to make a highly precise model for FLL, as its goal is to guide expression classification. Thus, we allow for small precision errors in the annotation of the ground truth landmark positions. Consequently, we choose Dlib's Ensemble Regression of Trees (ERT) model \cite{lr-ert} to obtain 68 landmark position ground-truth data. We use this data to train our network's landmark localization head, as shown in Fig. \ref{fig:combo_network_arch}. This ERT model has been trained on the iBUG 300-W face landmark \cite{lr-ibug}. 


\vspace{-0.5cm}

{\color{black}
\subsection{Neural Network Architecure}
\begin{figure*}[t]
\centering
\includegraphics[width=18cm,height=7cm]{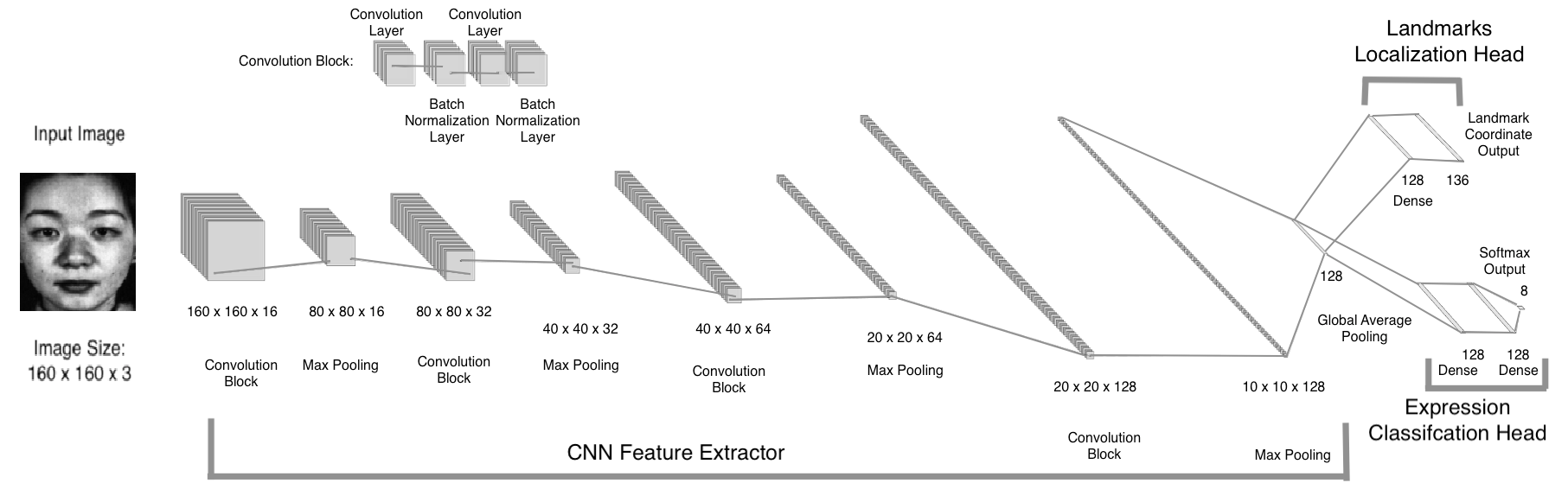}
\caption{Our FER network consists of three parts - Feature extractor, Landmarks Localization head, and Expression Classification Head. (Here, maximum landmark outputs vary from a minimum of 18  for 9 nose points to 136 for the entire 68 points, and the expression output size is 7 for JAFFE and SFEW and 8 for CK+ because of extra expression \emph{Contempt}).} 
\label{fig:combo_network_arch}
\end{figure*}
This section describes our proposed Part-based Ensemble Transfer Learning, the two baseline networks, and their components.

\subsubsection{Network Components}
Each of the FER networks mentioned above consists of two or all of these parts: CNN Feature Extractor, Landmarks Localization head, and Expression Classification head, as shown in Fig.~\ref{fig:combo_network_arch}. 

\textbf{CNN Feature Extractor} consists of a sequence of 4 Convolutional Blocks and Max Pooling layer pairs arranged alternatively, followed by a Global Average Pooling layer. Each convolutional block, in turn, consists of two convolution and two batch normalization layers arranged alternately. After each batch normalization layer, we apply the RELU \cite{tr-relu} activation. Batch normalization \cite{tr-batchnorm} enables faster and more stable training of the network. Further, we initialize each convolution layer using He uniform initializer \cite{tr-he}. The feature extractor's input is a grayscale face crop of 160 x 160 x 3, and the output is a 128-dimensional (128-D) feature vector. The feature extractor has about 0.29 M (million) parameters. Its architecture is the same across all three datasets and the three types of networks.

\textbf{Landmarks Localization Output Head} takes the 128-D feature vector as input and consists of two dense layers. We initialize each dense layer using the Glorot uniform initializer \cite{tr-glorot}. The first dense layer is 128-D in size, and the second is the output layer with a size that varies depending on the number of landmarks that need to be localized. Consequently, the number of localization head parameters varies from 0.019M for 18 neurons representing the 9 nose points to 0.025M for 136 neurons representing the entire 68 points. The architecture is the same across all three datasets and varies depending on the type of network.

\textbf{Expression Classification Output Head} takes the 128-D feature vector as input and consists of 3 dense layers. We initialize each dense layer using the Glorot uniform initializer. The first two are 128-dimensional layers, and the last one is an output layer of 7 or 8 neurons. JAFFE and SFEW datasets have labeled data for seven expressions, whereas the CK+ dataset has the expression label \emph{Contempt} and these seven expressions. The activation for the output neurons is softmax. The Classification head of the models trained on JAFFE, SFEW, and CK+ have approximately 0.035M parameters (maximum in CK+ because it has eight output neurons).  Its architecture is the same across the three types of networks but varies depending on the dataset.

\subsubsection{Network Types}
This section describes the architecture of Baseline, Full Transfer Learning, and Part-based Ensemble Transfer Learning networks. 

\textbf{Baseline Network}
The Baseline network comprises of CNN feature extractor and the classification head described above. An ensemble of 5 randomly initialized Baseline networks is trained for expression prediction.

\textbf{Full Transfer Learning Network}
The Full Transfer Learning (FTL) network consists of a CNN feature extractor, a Landmark localization head, and a classification head. Initially, we train the CNN feature extractor and localization head together to predict all 68 fiducial points at once. The output layer of the localization head has 136 neurons corresponding to 68 points. Then, we replace the localization head with the classification head described above and fine-tune the entire network for expression classification. Like the Baseline network, we employ an ensemble of 5 Full Transfer Learning networks to predict expression class.

\textbf{Part-based Ensemble Transfer Learning Network}
The proposed Part-based Ensemble Transfer Learning (EL) network comprises five transfer learning networks, such that each one focuses on one of the following facial features: eyebrows, eyes, nose, mouth, and jaw. Here, the jaw sub-network represents three regions of the face: jaw, chin, and cheek. During the landmark localization training phase, we divide these 68 landmarks into five subsets of points (as shown in Fig. \ref{fig:combo_all_lr}) and train an FLL model for each feature: eyebrows, eyes, nose, mouth, and jaw. The outputs of these five models are; 20 neurons representing the 10 eyebrows points, 24 neurons representing the 12 eyes points, 18 neurons representing the 9 nose points, 40 neurons representing the 20 mouth points, and 34 neurons representing the 17 jaw points. Like the Full Transfer Learning network, we replace the localization head of each sub-network with the classification head and fine-tune each sub-network for expression classification separately. Then, we predict the expression class using our ensemble policy. 

\subsection{Training process}
\label{training-nn}
Before training the models, we perform data augmentation like horizontal flipping, rotation, shear, and translation to increase the training data's size, as shown in Fig.~\ref{fig:pipeline_diagram} and normalize input between \(\left[-1,1\right]\). Additionally, for the transfer learning networks, the landmark coordinates undergo the same transformations as their corresponding images. After transformation, we normalize the landmark coordinates between \(\left[0,1\right]\). We have two different training strategies :

\subsubsection{Direct Expression Classification Training}
In this strategy, we train a network on the expression datasets without performing transfer learning and optimize the categorical crossentropy loss function described in Eq \ref{eq:ce}. We use an initial learning rate \(\alpha = 10^{-2}\) for training the model and stop training when the validation loss saturates. This strategy is only followed for the Baseline network. 

\subsubsection{Transfer Learning Expression Classification Training}
\label{sec:classes}
In this strategy, we divide the training process into three stages - Facial Landmark Localization pre-training, Expression Classification Head training, and Expression Classification fine-tuning. This strategy is followed for the Full Transfer Learning and Part-based Ensemble Transfer Learning networks.

\textbf{Facial Landmark Localization Pre-training} is formulated as a regression problem where we minimize the L1 loss:  

\begin{equation}\label{eq:mae}
 L_{1}= \left|y_{i}-\hat{y_{i}}\right|
 \end{equation}
 
where $y_{i} \in {\left[0,1\right]}^{z}$ is the ground truth coordinate for the $\emph{i-th}$ sample and $\hat{y_{i}}\ \in {\left[0,1\right]}^{z}$ is the coordinate predicted by the network and $z \in \{18,20,24,34,40, 136\}$ represents the number of output neurons based on the landmarks model being trained.

We train the facial landmark localization head and the feature extractor, using an initial learning rate \(\alpha = 10^{-2}\). We use saturation of validation loss as the stopping criteria for training. 

\textbf{Expression Classification Head Training}
In this stage, we freeze the weights of the feature extractor, replace the localization head with the classification head, and train the classification model. Only classification head weights are updated during this training using an initial learning rate \(\alpha = 10^{-2}\). We formulate this training as a 7 or 8 class classification problem where we minimize the categorical crossentropy loss:

\begin{IEEEeqnarray}{C}\label{eq:ce}
Categorical \ Cross \ Entropy = -\sum_{i=1}^{C}y_ilog(\hat{y_i})
\end{IEEEeqnarray}
Where C is the number of classes, \(Classes = \{Angry, Contempt*, Disgust, Fear, Happy, Neutral, Sad, \\
Surprise\}\), (\emph{*Contempt only in CK+}),  $y_i$ is the true class,  and $\hat{y_i}$ is predicted class, which is obtained after softmax activation, refer to Eq. \ref{eq:softmax}.

Moreover, training is done for a few epochs, as the validation loss saturates quickly. This intermediate stage aims to initialize the weights of the classification head for expression recognition while giving it the context of facial feature representation learned during the localization stage.

\textbf{Expression Classification Fine-tuning}
In this final stage, we unfreeze the feature extractor weights and fine-tune the entire train classification network. We use the same categorical cross-entropy loss described above to optimize the weights of the network. We fine-tune the network with a minimal initial learning rate \(\alpha = 10^{-4}\) to prevent unlearning of the feature extractor weights resulting from drastic weight update steps. It helps align the weights of the facial feature extractor, which were learned during the localization stage, to the expression recognition task and simultaneously enhances the expression recognition ability of the classification head.

\subsubsection{Training Parameters}
We train five model ensembles for Baseline and FTL networks for an equitable comparison with our proposed ensemble. Furthermore, each model is trained on one dataset only. For instance, the SFEW dataset model has only seen the SFEW training dataset's images. We do not pre-train our feature extractors on any other FER or face dataset. Model hyperparameter values are kept consistent across datasets and network types. We use Adam optimizer \cite{tr-adam} \( \{\beta_{1} = 0.9, \beta_{2} = 0.999, \epsilon = 10^{-7}\}\) and a mini-batch size of 32 images for training models. Further, the validation loss does not saturate after a fixed number of epochs. It varies depending on the dataset used for training the model. Following is the approximate number of epochs for each dataset and network type:

\textbf{CK+}
We train each baseline model for 400 epochs. We train the feature extractor and the localization head of the FTL and EL for 100 epochs. Then, we train the classification head and fine-tune the feature extractor for 300 epochs.

\textbf{JAFFE}
We train the Baseline network for 300 epochs. We train the feature extractor and the localization head of the FTL and EL networks for 100 epochs. Then, we train the classification head and fine-tune the feature extractor for 200 epochs.

\textbf{SFEW}
We train the Baseline network for 400 epochs. We train the feature extractor and the localization head of the FTL and EL networks for 200 epochs. Then, we train the classification head and fine-tune the feature extractor for 200 epochs.

Lastly, we use Tensorflow and Keras deep learning framework to train our models on a single Nvidia Tesla K80 GPU.

\subsection{Inferencing}
\label{sec:inferencing}
We fix the model weights learned during training and forward propagate test images through the model for inferencing expression and measuring model accuracy. The same ensemble policy, as described below, is employed for the Part-based Ensemble Transfer Learning network and the two baseline networks.

\textbf{Ensemble Prediction Policy} Each sub-network outputs a score vector of length C (number of classes), whose values correspond to the scores of the possible expression classes mentioned in Section \ref{sec:classes}. These scores are squashed to probability values between 0 and 1 using a softmax function described below.

\begin{equation}\label{eq:softmax}
\textnormal{Softmax}\left(s_{p}\right) = \frac{e^{s_{p}}}{\sum_{j}^{C} e^{s_{j}}}
\end{equation}
Where \(s_{p}\) is the score of the positive class, \(s_{j}\) is the score of the class \emph{j}. 
Then, we perform an expression-wise summation of these probabilities to get a final output vector and use it for prediction, as shown below:
\begin{equation}\label{eq:inference}
Predicted \ Label_{(ensemble)} = \operatorname*{arg\,max}_{k}\sum_{m} {Prob(s)}
\end{equation}
Where \(\ m \in \{eyebrows, eyes, nose, mouth, jaw\}\) model, s stands for score, \(k \in  Classes,\ Prob(s) \in \left[0,1\right]^{C}\) \\
}

\begin{table*}[!t]
\caption{Accuracy of expression classification of the Baseline, FTL: Full Transfer Learning and EL: Part-based Ensemble Transfer Learning networks.}{\vspace{-.2cm}}
\begin{center}
\renewcommand{\arraystretch}{1.2}
\renewcommand{\tabcolsep}{1.6mm}
\begin{tabular}{| c | c | c | c | c | c | c | c | c |}
    \toprule
    \multirow{3}{*}{Dataset} &
    \multicolumn{8}{c|}{Classification Accuracies (\%) of Neural Networks}
    \\
    \cline{2-9}
    &  \multirow{2}{*}{Baseline} & \multirow{2}{*}{FTL} & \multicolumn{5}{c|}{Individual EL} &  \multirow{2}{*}{EL}
    \\
    \cline{4-8}
    & & & Eyebrows & Eyes & Nose & Mouth & Jaw & \\
    \midrule
      CK+ & 82.56 & 95.87 & 93.29 & 93.96 & 91.86 & 95.16 & 93.28 & \textbf{97.31} \\
     JAFFE & 85.02 & 92.42 & 91.43 & 87.30 & 82.88 & 92.36 & 86.40 & \textbf{97.14} \\
      SFEW & 36.47 & 42.20 & 40.60 & 41.30 & 36.70 & 41.74 & 40.82 & \textbf{44.50} \\
    \bottomrule
\end{tabular}
\end{center}
\vspace{-0.4cm}
\label{tab:exp1-1}
\end{table*}

\begin{table}[!t]
\caption{Comparison of EL: Part-based Ensemble Transfer Learning Network with CB: Current Benchmark on the basis of Classification Accuracy and Model Parameters.}{\vspace{-.2cm}}
\begin{center}
\renewcommand{\arraystretch}{1.2}
\renewcommand{\tabcolsep}{1.6mm}
\begin{tabular}{| c | c | c | c | c |}
    \toprule
    \multirow{2}{*}{Dataset} &
    \multicolumn{2}{c|}{Classification Accuracies (\%)} &
    \multicolumn{2}{c|}{Model Parameters (M)}
    \\
    \cline{2-5}& CB & EL & CB & EL \\
    \midrule
      CK+ & 96.80 \cite{ckplus-dsp-5} & \textbf{97.31} & 11 & \textbf{1.65} \\
      JAFFE & 91.80 \cite{ckplus-dsp-1, dsp-paper} & \textbf{97.14} & 2 & \textbf{1.65} \\
      SFEW & \textbf{48.19} \cite{ckplus-dsp-5} & 44.50 & 11 & \textbf{1.65} \\
    \bottomrule
\end{tabular}
\end{center}
\vspace{-0.5cm}

\label{tab:exp1-2}
\end{table}

{\color{black}
\section{Experimental Setup \& Results}

\subsection{Benchmark Dataset Analysis}
In this section, we report the average expression classification accuracy of the Part-based Ensemble Transfer Learning, Full Transfer Learning, and Baseline networks for the three datasets - CK+, JAFFE, and SFEW (as shown in Table \ref{tab:exp1-1}) and compare them with the current benchmark (as shown in Table \ref{tab:exp1-2}). We also state the accuracy of each sub-network of the proposed ensemble when it is used independently. Moreover, in Table \ref{tab:exp1-2}, we compare our proposed ensemble network's accuracy and model parameter count with the current benchmark for the three datasets, whose values are taken from the respective papers. As the highest decimal precision of the reported accuracy values for the current benchmark results across all three datasets is two decimal places, we round off our accuracy values accordingly. 

We only present the benchmark results for each dataset in Table \ref{tab:exp1-2} for the sake of conciseness and clarity. Nevertheless, we compare our network's performance against other comparable research work. By comparable, we mean prior work which has the following characteristics: same version of the dataset as us (for example, SFEW 2.0 and not SFEW 1.0 dataset); the same evaluation policy (10-fold subject independent evaluation for CK+ and JAFFE dataset, and validation dataset evaluation for SFEW) as us; no pre-training on any other FER or face dataset; evaluating the result for seven expression classes for JAFFE and SFEW, and eight expression classes for CK+ datasets. Thus, we compare our results with the following prior works: \cite{ckplus-dsp-5, ckplus-dsp-2, ckplus-dsp-4} for CK+, \cite{ckplus-dsp-1,jaffe-1-1, jaffe-1-2, jaffe-1-3, jaffe-1, jaffe-1-4} for JAFFE, and \cite{ckplus-dsp-5,ckplus-dsp-2,sfew-dsp-6,pw-dsp-15,sfew-dsp-7} for SFEW. Lastly, we also illustrate confusion matrices to analyze EL's expression-wise accuracy. For CK+ and JAFFE (Fig. \ref{fig:cf_ckplus} and Fig. \ref{fig:cf_jaffe} respectively), the confusion matrix is the summation of the ten confusion matrices, one for each subject independent cross-validation fold. In the case of SFEW, we construct the confusion matrix by evaluating our proposed ensemble network on the SFEW validation dataset, shown in Fig. \ref{fig:cf_sfew}.

\begin{figure}[!h]
\centering
\newcommand\items{8} 
\arrayrulecolor{white} 
\noindent\begin{tabular}{cc*{\items}{|E}|}
\multicolumn{1}{c}{} &\multicolumn{1}{c}{} &\multicolumn{\items}{c}{Predicted} \\ \hhline{~*\items{|-}|}
\multicolumn{1}{|c|}{} & 
\multicolumn{1}{c}{} & 
\multicolumn{1}{c}{\rot{Angry}} & 
\multicolumn{1}{c}{\rot{Contempt}} & 
\multicolumn{1}{c}{\rot{Disgust}} & 
\multicolumn{1}{c}{\rot{Fear}} &
\multicolumn{1}{c}{\rot{Happy}} & 
\multicolumn{1}{c}{\rot{Neutral}} &
\multicolumn{1}{c}{\rot{Sad}} & 
\multicolumn{1}{c}{\rot{Surprise}} 

\\ \hhline{~*\items{|-}|}
\multirow{\items}{*}{\rotatebox{90}{Actual}} 
&Angry& 132 & 0 & 0 & 0 &  0 & 3 & 0 & 0 \\ \hhline{~*\items{|-}|}
&Contempt& 0 & 48 & 0 & 0 &  0 & 6 & 0 & 0 \\ \hhline{~*\items{|-}|}
&Disgust& 0 & 0 & 177 & 0 & 0 &  0 & 0 & 0 \\ \hhline{~*\items{|-}|}
&Fear& 0 & 0& 0 & 66 & 0 &  3 & 3 & 3  \\ \hhline{~*\items{|-}|}
&Happy& 0 & 0& 0 & 0 & 207 &  0 & 0 & 0 \\ \hhline{~*\items{|-}|}
&Neutral& 1 & 0& 0 & 0 & 0 & 326 & 0 & 0 \\ \hhline{~*\items{|-}|}
&Sad& 0 & 0 & 0 & 0 &  0 & 10 & 74 & 0 \\ \hhline{~*\items{|-}|}
&Surprise& 0 & 3 & 1 & 0 & 0 &  3 & 0& 242  \\ \hhline{~*\items{|-}|}

\end{tabular}
\caption{Confusion matrix for the proposed ensemble network's results on all the ten validation folds of the CK+ dataset.}
\label{fig:cf_ckplus}
\end{figure}

\begin{figure}[!h]
\centering
\newcommand\items{7} 
\arrayrulecolor{white} 
\noindent\begin{tabular}{cc*{\items}{|E}|}
\multicolumn{1}{c}{} &\multicolumn{1}{c}{} &\multicolumn{\items}{c}{Predicted} \\ \hhline{~*\items{|-}|}
\multicolumn{1}{|c|}{} & 
\multicolumn{1}{c}{} & 
\multicolumn{1}{c}{\rot{Angry}} & 
\multicolumn{1}{c}{\rot{Disgust}} & 
\multicolumn{1}{c}{\rot{Fear}} &
\multicolumn{1}{c}{\rot{Happy}} & 
\multicolumn{1}{c}{\rot{Neutral}} &
\multicolumn{1}{c}{\rot{Sad}} & 
\multicolumn{1}{c}{\rot{Surprise}} 

\\ \hhline{~*\items{|-}|}
\multirow{\items}{*}{\rotatebox{90}{Actual}} 
&Angry& 29 & 0 & 0 & 0 &  0 & 1 & 0 \\ \hhline{~*\items{|-}|}
&Disgust& 0 & 29 & 0 & 0 &  0 & 0 & 0 \\ \hhline{~*\items{|-}|}
&Fear& 0 & 0& 32 & 0 &  0 & 0 & 0  \\ \hhline{~*\items{|-}|}
&Happy& 0 & 0& 0 & 30 &  0 & 0 & 1 \\ \hhline{~*\items{|-}|}
&Neutral& 0 & 0& 0 & 0 & 30 & 1 & 0 \\ \hhline{~*\items{|-}|}
&Sad& 0 & 1 & 0 & 1 &  0 & 29 & 0 \\ \hhline{~*\items{|-}|}
&Surprise& 0 & 0& 0 & 1 &  0 & 0& 29  \\ \hhline{~*\items{|-}|}

\end{tabular}
\caption{Confusion matrix for the proposed ensemble network's results on all the ten validation folds of the JAFFE dataset.}
\label{fig:cf_jaffe}
\end{figure}

\begin{figure}[!h]
\centering
\newcommand\items{7} 
\arrayrulecolor{white} 
\noindent\begin{tabular}{cc*{\items}{|E}|}
\multicolumn{1}{c}{} &\multicolumn{1}{c}{} &\multicolumn{\items}{c}{Predicted} \\ \hhline{~*\items{|-}|}
\multicolumn{1}{c}{} & 
\multicolumn{1}{c}{} & 
\multicolumn{1}{c}{\rot{Angry}} & 
\multicolumn{1}{c}{\rot{Disgust}} & 
\multicolumn{1}{c}{\rot{Fear}} &
\multicolumn{1}{c}{\rot{Happy}} & 
\multicolumn{1}{c}{\rot{Neutral}} &
\multicolumn{1}{c}{\rot{Sad}} & 
\multicolumn{1}{c}{\rot{Surprise}} 

\\ \hhline{~*\items{|-}|}
\multirow{\items}{*}{\rotatebox{90}{Actual}} 
&Angry& 32 & 3 & 3 & 9 &  13 & 7 & 10 \\ \hhline{~*\items{|-}|}
&Disgust& 3 & 1 & 1 & 5 &  6 & 5 & 2 \\ \hhline{~*\items{|-}|}
&Fear& 13 & 2 & 1 & 7 &  9 & 10 & 5  \\ \hhline{~*\items{|-}|}
&Happy& 3 & 0 & 1 & 54 &  4 & 10 & 1 \\ \hhline{~*\items{|-}|}
&Neutral& 5 & 2& 3 & 2 & 56 & 15 & 3 \\ \hhline{~*\items{|-}|}
&Sad& 0 & 0 & 2 & 3 &  19 & 41 & 8 \\ \hhline{~*\items{|-}|}
&Surprise& 7 & 0& 0 & 6 &  26 & 9 & 9  \\ \hhline{~*\items{|-}|}

\end{tabular}
\caption{Confusion matrix for the proposed ensemble network's results on the validation set of the SFEW dataset.}
\label{fig:cf_sfew}
\end{figure}

\begin{table*}[!t]
\caption{Cross-Dataset Generalization test of the Baseline, FTL: Full Transfer Learning and EL: Part-based Ensemble Transfer Learning networks from SFEW Train dataset to CK+ and JAFFE datasets.}{\vspace{-.2cm}}
\arrayrulecolor{black}
\begin{center}
\renewcommand{\arraystretch}{1.2}
\renewcommand{\tabcolsep}{1.6mm}
\begin{tabular}{| c | c | c | c | c | c | c | c | c | c |}
    \toprule
    \multirow{3}{*}{Train Dataset} &
    \multirow{3}{*}{Test Dataset} &
    \multicolumn{8}{c|}{Classification Accuracies (\%) of Neural Models}
    \\
    \cline{3-10}
    &  &
    \multirow{2}{*}{Baseline} & \multirow{2}{*}{FTL} & \multicolumn{5}{c|}{Individual EL} &  \multirow{2}{*}{EL}
    \\
    \cline{5-9} & & & & Eyebrows & Eyes & Nose & Mouth & Jaw & \\
    \midrule
      SFEW Train & SFEW Valid & 36.47 & 42.20 & 40.60 & 41.30 & 36.70 & 41.74 & 40.82 & 44.50 \\
      SFEW Train & CK+ & 32.85 & 41.31 & 33.33 & 35.49 & 22.97 & 37.72 & 32.62 & \textbf{47.53} \\
      SFEW Train & JAFFE & 16.90 & 26.29 & 22.54 & 24.88 & 18.78 & 24.88 & 21.60 & 30.51 \\
    \bottomrule
\end{tabular}
\end{center}
\label{tab:exp2}
\end{table*}

\subsection{Study to evaluate the usefulness of transfer learning from Facial Landmark Localization to Facial Expression Recognition}
Our proposed network's design and training strategy are based on the premise that FLL and FER share similarities, and transfer learning from the facial representations learned from the former can enhance the latter's performance. To verify this premise, we compare the fiducial points predicted before and after fine-tuning the feature extractor for expression classification, keeping the landmark regressor constant. We use the SFEW dataset for this experiment as its images can be taken as a good approximation of faces a model will encounter in the real world. Before fine-tuning, we use sub-networks of the ensemble, which have only been trained for facial landmark localization, for predicting the corresponding set of fiducial points, as shown in Fig \ref{fig:retain}. After fine-tuning, we replace the classification head with the previously trained localization head for each sub-network (architecture shown in Fig \ref{fig:combo_network_arch}). Without re-tuning the weights of the feature extractor and the localization head, we use it to predict the fiducial points, as shown in Fig \ref{fig:retain}. Then, we compare the mean absolute difference (Eq. \ref{eq:mae}) between predicted fiducial points, both before and after fine-tuning, with the ground truth points. We also visually evaluate the similarity between the points predicted before and after fine-tuning.

\subsection{Visual Dissection Study}

To interpret how the different convolutional sub-networks of the ensemble network work, we employ Gradient-weighted Class Activation Mapping (Grad-CAM) \cite{vis-gradcam}. Grad-CAM uses the gradient information flowing into the last convolutional layer of CNN to understand each neuron for a class label of interest. We obtain the class discriminative localization map of width u and height v for any expression class c by first computing the gradient of the score for that class, that is, \({y}^{c}\) (before the softmax), for feature maps \(\alpha_{k}\) of a convolutional layer. These gradients flowing back are global average-pooled over the width and height dimensions (indexed by i and j respectively) to obtain the neuron importance weights \(\alpha_{k}^{c}\). 

\begin{equation}\label{eq:gradcam1}
\alpha_{k}^{c}=\overbrace{\frac{1}{Z} \sum_{i} \sum_{j}}^{\text {global\ average\ pooling }} \underbrace{\frac{\partial y^{c}}{\partial A_{i j}^{k}}}_{\text {gradients\ via\ backprop }}
\end{equation}

After calculating \(\alpha_{k}^{c}\), we perform a weighted combination of the activation maps and follow it by a ReLU. Without it, the class activation map highlights more than required and achieves low localization performance. 

\begin{equation}\label{eq:gradcam2}
L_{\mathrm{Grad}-\mathrm{CAM}}^{c}=\operatorname{ReLU} \underbrace{\left(\sum_{k} \alpha_{k}^{c} A^{k}\right)}_{\text {linear \ combination }}
\end{equation}

Subsequently, we superimpose the activation map (heatmap) with the original image to coarsely visualize which region of the face each sub-network of our proposed ensemble focuses on while performing FER. As we use a 5-model ensemble for the FTL and Baseline networks, their final heatmaps (as shown in Fig. \ref{fig:gradcam}) are the result of the union of 5 heatmaps. Subsequently, the final heatmap is superimposed on the image to get the overlap map. 

\subsection{Cross-Dataset Generalization}

In cross-dataset generalization tests, we estimate the expression recognition accuracy of the model trained and tested on two datasets having different data characteristics and expression-wise distribution. In terms of accuracy, the degree of cross-dataset generalizability of the model depends on how well it classifies on a dataset on which it was not trained, as compared to its test accuracy on the dataset on which it was trained. Additionally, we also evaluate our model's generalizability on faces in real-world unconstrained settings through this test. We know that the SFEW dataset consists of facial images with varied poses, sizes, and illumination. Whereas the lab-controlled CK+ and JAFFE datasets consist of fewer subjects and consist of only faces with frontal poses, thereby representing an ideal condition for facial expression analysis, not a generalized one. 

Therefore, we perform a Cross-Dataset generalization test, where we use the SFEW train dataset as the training set and CK+ and JAFFE datasets for testing. While testing on the CK+ dataset, we drop the samples whose expression class is Contempt as this class is absent in the SFEW dataset. We perform this test for each network type - EL, FTL, and Baseline. We also report the independent generalization ability of each sub-network of the proposed ensemble, as shown in Table \ref{tab:exp2}. 

\subsection{Computational Complexity}
In this section, we report hardware-independent and deep learning framework agnostic model complexity in terms of FLOPs for inference and memory for storing parameters. To report the worst-case complexities, we utilize the proposed ensemble models trained on the CK+ dataset, as they have the maximum number of parameters among the models trained on the three datasets. A sub-network of the ensemble transfer learning network has about \textbf{0.33M} parameters in the worst case (8 outputs of CK+ dataset). As the ensemble transfer learning network comprises five such models, it has about \textbf{1.65M} parameters. 

To calculate the FLOPs for inference, we sum up the number of multiplication and addition operations for each network layer. We only consider the convolution, fully connected (or dense), and max-pooling layers for calculation purposes. The operations contributed by the activation function, and the global average pooling layers are extremely small compared to the layers mentioned above and can be neglected. Further, the batch norm layer's operations are folded into the weights of the preceding convolution layer, which is possible as the batch norm layer directly succeeds the convolution layer in our architecture and consists of linear transformations like the convolution layer. We use the following formulas to measure FLOPs for the three different types of layers.

\begin{flalign}\label{eq:flops-conv}
\textnormal{FLOPs Convolution layer} =2 \times C_{i} \times k^{2} \times C_{o} \times W \times H &&
\end{flalign} 

\begin{flalign} \label{eq:flops-dense}
\textnormal{FLOPs Fully Connected layer} = 2 \times I \times O &&
\end{flalign} 


\begin{flalign}\label{eq:flops-maxpool}
\textnormal{FLOPs Max Pooling layer} = \frac{W}{stride} \times \frac{H}{stride} \times C_{o} &&
\end{flalign} 

Where \(C_{i}\) is the number of channels of the input feature map (or image), k is the size of kernel or filter, \(C_{i}\) is the number of channels of the output feature map, W is the width of the input feature map (or image), H is the height of the input feature map (or image), I is the number of input neurons to a fully connected layer, and O is the number of output neurons of the same fully connected layer, stride is the max pool stride length.

We obtain that each sub-network requires a total of approximately 0.555 GFLOPs of inference, which is in the range of FLOPs of MobileNet \cite{mbv1,mbv2}, a neural network designed for low latency and fast inference. In total, the ensemble network requires 2.775 GFLOPs for inference. Further, the convolutional layers contribute to 99.98\% of the total FLOPs, due to which we can employ parallelized CNN implementation for even faster inference. 

To calculate the memory required to store model parameters, we multiply the total number of parameters in a subnetwork by 4, as the weights are stored in float32 format, which occupies 4 bytes of space. We obtain that each subnetwork and the ensemble of five sub-networks approximately occupy \textbf{6.3 MB} and \textbf{31.5 MB} of space, respectively. 

\section{Discussion}
\subsection{Benchmark Dataset Analysis}
From Table \ref{tab:exp1-1}, we can infer that the Full Transfer Learning network outperforms the Baseline network across all three datasets, thereby showing the benefit of transfer learning from Facial Landmark Localization over direct expression classification training. Further, the part-based ensemble transfer learning method outperforms the 5-model ensembles of the FTL and Baseline networks, thereby demonstrating the merit of the part-based ensemble approach. 

Based on our ensemble policy \ref{eq:inference}, if the sub-models of the proposed network were redundant, they were more likely to produce an average accuracy of all the sub-models or, in the best-case an accuracy as high as the highest accuracy achieved by one of the sub-models. However, we observe that the accuracy of the proposed ensemble is 4.7\% higher than the average accuracy of the five sub-models and 2.76\% higher than the mouth sub-model (highest accuracy among the sub-models). Thus, we attribute this observation to the complementary nature of the sub-networks, a key design parameter of our proposed ensemble.

We observe that the sub-network focussing on the mouth region has the highest independent classification accuracy. After the mouth region, eyes and eyebrows have the second-highest independent classification accuracy, which is further followed by jaw and nose (with the nose having the least). Now, we try to relate these findings with the Facial Action Coding System (FACS) Main Action units (AUs) and how they are combined to describe emotions - anger, disgust, fear, happiness, sadness, and surprise. We find that action units corresponding to the mouth region (lip movement) produce the most diverse visual variations. Further, the lip AUs used to describe the expressions are often unique and specific to an expression. Then, we have the AUs describing the eyes (eyelid movement) and eyebrows (brow movement), which produce the second most visual variations. Unlike lip AUs, eyelid and brow AUs are often the same across different expressions. We consider cheek, chin, and jaw AUs together to draw correspondence with our proposed ensemble's jaw sub-network. Like eyelid and brow, these AUs are often the same across expressions, but they are used in fewer expressions than even eyelid and brow AUs. Lastly, only one nose AU is used to describe an expression, which is disgust. As expression classification from static images is a pattern recognition problem, higher discriminability of patterns corresponds to higher classification accuracy. From the above discussion on AUs, we obtain the following order of discriminability - mouth, eyes, eyebrows, jaw, and nose. This order conforms with the results we have obtained (shown in Table \ref{tab:exp1-1}), thereby demonstrating that our model can correlate the visual patterns emanating from facial muscles' motor movements to emotions, a key design parameter for our network.

On comparing our proposed part-based ensemble's results with the three datasets' current benchmarks, we observe that our proposed ensemble obtains the highest reported accuracy of \textbf{97.31\%} for CK+ and \textbf{97.14\%} for JAFFE (subject to evaluation criteria), outperforming the benchmark for CK+ by \textbf{0.51 \%}, and Jaffe by \textbf{5.34\%}, as shown in Table \ref{tab:exp1-2}. It falls short of the benchmark for SFEW by 3.69 \%, but has nearly 1/7th the total model parameters compared to the benchmark network. Moreover, our network consistently has a lesser number of total model parameters than the benchmarks. We know that training time and storage space depend on model (trainable + non-trainable) parameters. This means that our proposed ensemble model, which has lesser model parameters than the current benchmarks, is more computationally efficient. Thus, our part-based ensemble transfer learning network performs a good trade-off between accuracy and computational efficiency, another key design parameter for our network. 

Furthermore, the confusion matrices reveal information about the underlying expression-wise distribution of the three datasets. CK+ and SFEW datasets have an imbalanced expression-wise distribution. This imbalance is evident in Fig. \ref{fig:cf_ckplus} and in Fig. \ref{fig:cf_sfew}, where the maximum false positives are for the class neutral, which happens to be the most prevalent class in both the datasets. On the contrary, the JAFFE dataset, a nearly balanced expression-wise dataset, has no such prediction bias for a single expression class (shown in Fig. \ref{fig:cf_jaffe}).

\begin{figure}[!h]
	\begin{minipage}{\columnwidth}
	    \centering  
        \begin{subfigure}[t]{.30\textwidth}
            \centering
    		\includegraphics[width=2.5cm,height=5cm]{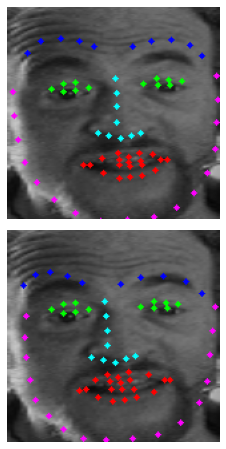}
	    \end{subfigure}
	    \hspace{0.15cm}
	    \begin{subfigure}[t]{.30\textwidth}
            \centering
            \includegraphics[width=2.5cm,height=5cm]{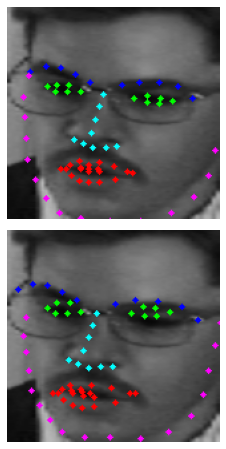}
	    \end{subfigure}
	    \hspace{0.15cm}
	    \begin{subfigure}[t]{.30\textwidth}
        \centering
            \includegraphics[width=2.5cm,height=5cm]{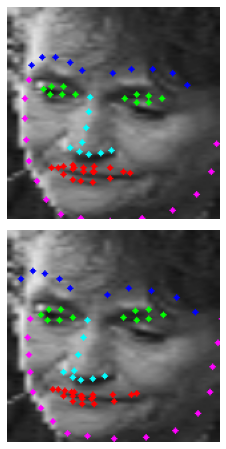}
	    \end{subfigure}%
	\end{minipage}

	\begin{minipage}{\columnwidth}
		\vspace{0.25cm}
	    \centering  
        \begin{subfigure}[t]{.30\textwidth}
            \centering
    		\includegraphics[width=2.5cm,height=5cm]{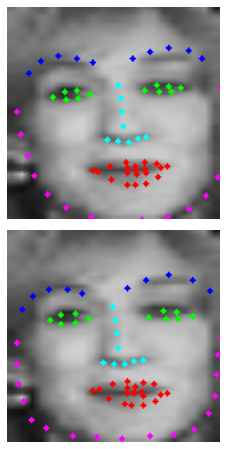}
	    \end{subfigure}
	    \hspace{0.15cm}
	    \begin{subfigure}[t]{.30\textwidth}
            \centering
            \includegraphics[width=2.5cm,height=5cm]{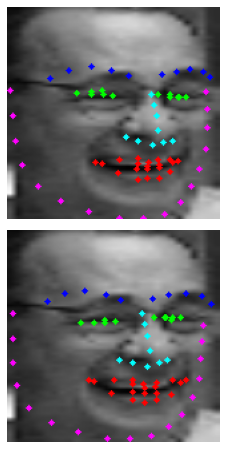}
	    \end{subfigure}
	    \hspace{0.15cm}
	    \begin{subfigure}[t]{.30\textwidth}
        \centering
            \includegraphics[width=2.5cm,height=5cm]{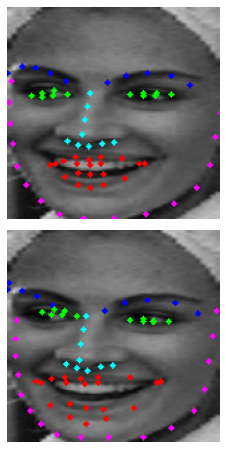}
	    \end{subfigure}%
	\end{minipage}

	    	\caption{{\color{black}In each of the six pairs of faces, the top one represents fiducial points predicted before fine-tuning the feature extractor for expression classification, and the bottom one represents points predicted after fine-tuning while keeping landmark localization head weights constant. Here the images belong to the SFEW dataset, and the points corresponding to eyebrows are represented by blue, eyes by green, nose by light blue, mouth by red, and jaw by pink color.}}
	    		\label{fig:retain}
\end{figure}

\subsection{Analyzing model's ability to perform Facial Landmark Localization after Expression Classification fine-tuning}
\label{sec:relearn}
We obtain that the mean absolute difference between predicted and growth truth points has decreased for mouth and eyes and remained almost similar for jaw post fine-tuning. On the other hand, a slight increase in absolute difference is noticed for eyebrows and a comparatively larger difference for nose. Interestingly, the points on the mouth capture its orientation better after fine-tuning, which also corroborates the reduction in its mean absolute difference. Similar visual correlations with reduction or increase in mean absolute difference have been found for other facial features. Moreover, from Table \ref{tab:exp1-1}, we observe a similarity in the classification accuracies of each sub-network and the corresponding change in mean absolute difference and the spatial orientation captured by the fiducial points after fine-tuning. 

Further, we consider that two networks with the same architecture will produce similar facial landmark localization when the base CNN extracts similar features and the regressor heads are equally good at using these features to predict the points. As the regressor head is constant before and after fine-tuning, the CNN base feature extractor must extract similar facial representation to exhibit similar localization ability. From experimental results, we obtain that the facial representation capabilities before and after supervised fine-tuning on two different sets of loss functions and labels (tasks) are similar. Thus, from the argument mentioned above and experimental observations, we can safely claim that Facial Landmark Localization and Facial Expression Recognition tasks share similarities. It also demonstrates that our training strategy helped our proposed ensemble learn nearly task-invariant features, an essential requirement for effective transfer learning. 

Another exciting outcome of this experiment is the possibility of cyclic consistency between these two tasks, which is evident from the results for the mouth sub-network learning. This property will need to be analyzed through more experimentation, which we plan to undertake as part of our future work.

\subsection{Explaining expression prediction through Grad-CAM Visualization}

The heatmap and overlap maps for the Baseline network in Fig \ref{fig:gradcam} show that the network does not focus on facial features specifically. Instead, it is trying to find a pattern from the entire face. On the other hand, we observe that in Fig \ref{fig:combo_ckplus_final} FTL network focuses on the eyes, eyebrows, and jaw to detect the expression of surprise; in Fig \ref{fig:combo_jaffe_final} FTL network focuses on the eyes and nose to detect the expression of disgust; in Fig \ref{fig:combo_sfew_final} FTL network focuses on the eyes and mouth to detect the expression happy. This demonstrates that transfer learning from the FLL task enables the FTL network to focus on facial features for expression prediction. Lastly, we visualize our proposed ensemble network. From Fig. \ref{fig:combo_ckplus_final}, \ref{fig:combo_jaffe_final} and \ref{fig:combo_sfew_final} , we observe that each sub-network focuses on one facial feature to make its independent expression prediction, which is later combined using the ensemble prediction policy. These visualizations demonstrate the complementary nature of our ensemble's sub-networks, a critical parameter for a successful ensemble network.

\begin{figure*}[!t]
	\centering
	\begin{subfigure}{.3\textwidth}
        \centering
		\includegraphics[width=0.9\linewidth,height=7.5cm]{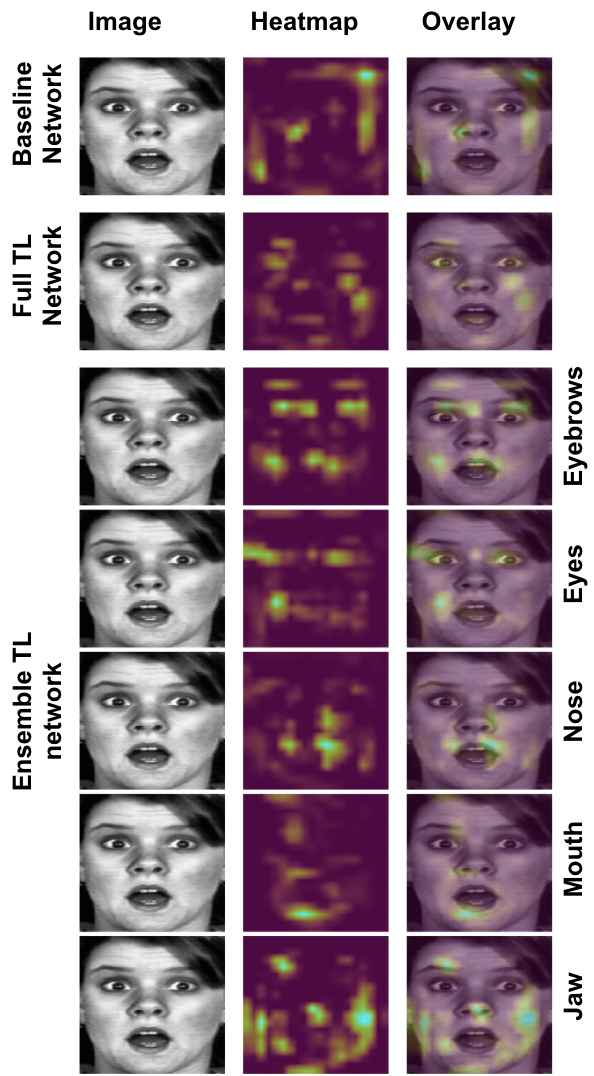}
        \caption{CK+ image with \emph{Surprise} expression.}
        \label{fig:combo_ckplus_final}
	\end{subfigure}
	\begin{subfigure}{.3\textwidth}
        \centering
        \includegraphics[width=0.9\linewidth,height=7.5cm]{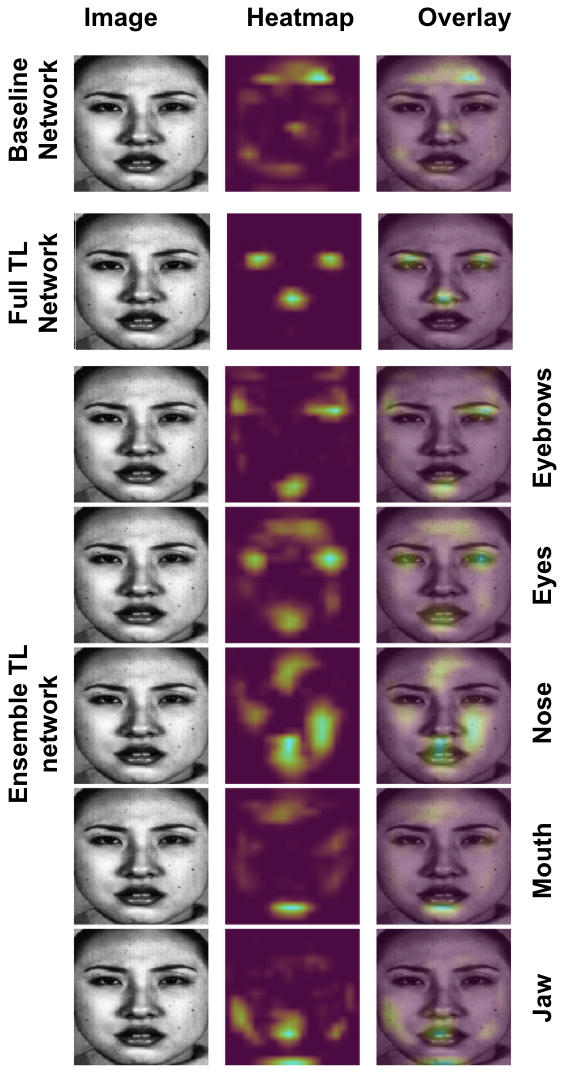}
        \caption{JAFFE image with \emph{Disgust} expression.}
		\label{fig:combo_jaffe_final}
	\end{subfigure}
	\begin{subfigure}{.3\textwidth}
        \centering
        \includegraphics[width=0.9\linewidth,height=7.5cm]{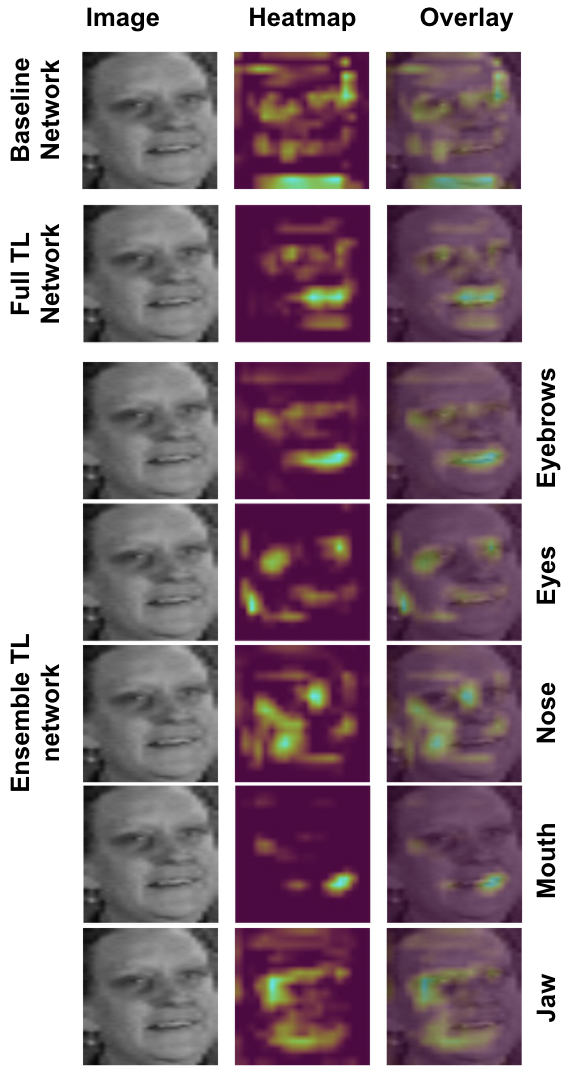}
        \caption{SFEW image with \emph{Happy} expression.}
		\label{fig:combo_sfew_final}
	\end{subfigure}%
	\caption{Grad-CAM visualization of the Baseline network, Full Transfer Learning network, and each sub-network of the Part-based Ensemble Transfer Learning network for a given expression.}
	\label{fig:gradcam}
\end{figure*}

\subsection{Cross-Dataset Generalization}
From Table \ref{tab:exp2}, we observe that the classification accuracy achieved by all three networks on the CK+ dataset is at par with those achieved on the SFEW Valid dataset. Moreover, the part-based ensemble achieves an accuracy of \textbf{47.53\%} on the CK+ dataset, which is even higher accuracy than the accuracy it obtains on the SFEW valid dataset. This shows that our proposed ensemble shows high generalization from SFEW to the CK+ dataset. On the other hand, the level of generalization from SFEW to JAFFE is comparatively less. This can be attributed to the dataset properties like gender and ethnicity covered in the datasets' images. Both SFEW and CK+ datasets consist of multiple subjects from different ethnic groups and gender. In contrast, the JAFFE dataset consists of only 10 Japanese female subjects, which imposes a large constraint on ethnicity and gender. This means that images in the JAFFE dataset will be equivalent to a small subset of the training images in the SFEW dataset. Consequently, a model mispredicting this small subset of SFEW images during training will contribute a small portion of the total training error. However, such a trained model will produce a large test error on a test set with the same distribution as the JAFFE dataset.

Lastly, we compare the generalizability of the part-based ensemble and our two baseline networks. We observe that the FTL network performs better than the Baseline network, thereby highlighting the advantage of transfer learning from FLL to FER. Further, our proposed ensemble outperforms the FTL network, demonstrating the benefit of a part-based ensemble over a 5-model ensemble of the FTL network.

\subsection{Limitation}
The proposed ensemble model falls short of the current benchmark on the SFEW dataset. We believe this is a limitation in learning an effective facial feature vector. The feature extractor is primarily trained during the FLL phase. Only fine-tuning with a small learning rate is applied during the classification phase; thus, the facial representation is majorly learned during the FLL phase. During the FLL phase, the feature extractor learns a representation to minimize the error in predicting landmark positions. Section \ref{sec:relearn} highlights that facial representation capabilities of our proposed ensemble remain nearly consistent before and after fine-tuning the feature extractor for expression classification. So, in the case of significant errors in ground truth landmark positions, noise is likely to be introduced in the representation learned by the feature extractor. 

Consequently, we visualized the ERT model used in our FER pipeline and observed that it made large mistakes in predicting landmark positions for SFEW images with extreme poses up to 90$^{\circ}$, as shown in Fig.~\ref{fig:landmark_issue}. The misalignment of ground truth landmark positions with corresponding facial parts most likely reduced the quality of facial representations learned by our proposed network, thereby reducing its expression prediction ability.

\begin{figure}[t]
\centering
\includegraphics[width=7cm,height=2cm]{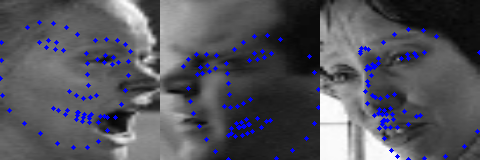}
\caption{Landmarks predicted by the ERT algorithm on the contrast-enhanced face crops are used as input to the neural network. Each image is taken from the train set of the SFEW dataset and has a high magnitude of head pose angle.}
\label{fig:landmark_issue}
\vspace{-2.5mm}
\end{figure}

\section{Conclusion}

In this work, we propose a Part-based Ensemble Transfer Learning network, which models how humans recognize facial expressions. We show that our proposed ensemble network uses visual patterns emanating from facial muscles' motor movements to predict expressions. We demonstrate the similarity between facial landmark localization to facial expression recognition and the usefulness of transfer learning from the former to the latter. The proposed network achieves ceiling level accuracy of \textbf{97.31\%} on the CK+ dataset and \textbf{97.14\%} on the JAFFE dataset (for the chosen evaluation policy) and outperforms the benchmark by \textbf{0.51\%} and \textbf{5.34\%} respectively. Across all three datasets, we observe that transfer learning from Facial Landmark Localization to Facial Expression recognition leads to better classification accuracy than directly training the network to classify expression. This strengthens our premise of using transfer learning from FLL to FER. 

Moreover, Grad-CAM visualization shows that transfer learning helps the final classification model focus on the facial feature(s) for an accurate classification compared to the non-transfer learning baseline network. It also demonstrates that each sub-network of the part-based ensemble focuses on a specific facial feature for expression prediction. This helps validate the complementary nature of the sub-networks of the proposed ensemble. The proposed ensemble network approach is not only accurate but also computationally efficient. It has a total of \textbf{1.65M} model parameters that are lesser than the parameters of the current benchmark for all three datasets. It requires 2.775 GFLOPs for inference. The convolutional layers contribute to 99.98\% of the total FLOPs, due to which we can employ parallelized CNN implementation for even faster inference. Moreover, it requires only \textbf{31.5 MB} of storage space in the worst case.

The Cross-dataset generalization tests reveal that generalizability is the highest for our Part-based ensemble Transfer Learning network, amongst the three networks described in this paper. Our proposed ensemble trained on the SFEW Train dataset achieves a classification accuracy of \textbf{47.53\%} on the CK+ dataset, which is higher than what it achieves on the SFEW Valid dataset. The generalization accuracy on the JAFFE dataset is comparatively low, owing to the significant constraint this dataset imposes on ethnicity and gender. Overall the cross-dataset generalizability of our proposed ensemble is high, which makes it suitable for real-world usage.

As part of our continued work, we plan to experiment with more robust and accurate techniques for landmark ground-truth annotation. We intend to explore the option of pre-training our proposed ensemble on the FER2013  dataset \cite{fer-dataset-paper}, a method used by recent works \cite{sfew-dsp-1,sfew-dsp-2}. Due to pre-training, the neural networks in these works have achieved accuracies north of 50\% on the SFEW valid dataset. We also plan to evaluate the effect of pre-training on cross-dataset generalization tests. Additionally, we will test our approach on larger datasets like FER2013 and RAF-DB. Lastly, we intend to explore the possibility of cyclic consistency between Facial Landmark Localization and Facial Expression Recognition tasks.
}

\ifCLASSOPTIONcompsoc
  \section*{Acknowledgments}
\else
  \section*{Acknowledgment}
\fi

The work is supported by MeitY(Government of India) under grant 4(16)/2019-ITEA.

\ifCLASSOPTIONcaptionsoff
  \newpage
\fi

\bibliographystyle{IEEEtran}
\bibliography{paper.bib}

\end{document}